\documentclass[10pt, a4paper]{article}
\usepackage{lrec}
\usepackage{multibib}
\newcites{languageresource}{Language Resources}
\usepackage{graphicx}
\usepackage{tabularx}
\usepackage{soul,color}
\usepackage{multirow}
\usepackage{balance}
\usepackage{makecell}

\usepackage{lipsum}
\usepackage{epstopdf}
\usepackage[utf8]{inputenc}

\usepackage{hyperref}
\usepackage{xstring}
\usepackage{spverbatim} 

\title{A Large Parallel Corpus of Full-Text Scientific Articles}

\name{Felipe Soares, Viviane Pereira Moreira, Karin Becker}

\address{Institute of Informatics, Universidade Federal do Rio Grande do Sul \\
         Porto Alegre - RS, Brazil \\
         \{felipe.soares, viviane, karin.becker\}@inf.ufrgs.br\\}

\abstract{
The Scielo database is an important source of scientific information in Latin America, containing articles from several research domains. A striking characteristic of Scielo is that many of its full-text contents are presented in more than one language, thus being a potential source of parallel corpora. In this article, we present the development of a parallel corpus from Scielo in three languages: English, Portuguese, and Spanish. Sentences were automatically aligned using the Hunalign algorithm for all language pairs, and for a subset of trilingual articles also. We demonstrate the capabilities of our corpus by training a Statistical Machine Translation system (Moses) for each language pair, which outperformed related works on scientific articles. Sentence alignment was also manually evaluated, presenting an average of 98.8\% correctly aligned sentences across all languages. Our parallel corpus is freely available in the TMX format, with complementary information regarding article metadata.    \\ \newline 
 }

\begin{document}

\maketitleabstract

\section{Introduction}
Cross-language corpora is one of the basis of Statistical Machine Translation (SMT) systems. 
The acquisition of quality corpora is not a trivial task, as it may demand considerable use of expert human curating, especially for parallel corpora. 
In that sense, the automated building of parallel corpora from open resources is of great interest in Natural Language Processing (NLP). 
Europarl \cite{koehn2005europarl}, is one of the largest parallel corpora availabe, including up to 21 European languages. 
Similarly, the United Nations (UN) parallel corpus \cite{ZIEMSKI16.1195} makes use of UN's official documents and records, providing aligned sentenced in six languages. 
Other parallel corpora initiatives have been reported in distinct domains, such as patents \cite{utiyama2007japanese}, movie subtitles \cite{zhang2014dual}, and books \cite{skadicnvs2014billions}.

Parallel corpora based on scientific articles can be a valuable language resource. Several text mining tasks may benefit from the availability of parallel corpora of scientific articles. The most straightforward example is cross language plagiarism detection, when an original text is translated in another language and presented as novel. Other possible applications are related to article indexing or classification, as well as the development and extension of Named Entity Recognition (NER) tools for multiple languages. The latter is of particular interest to the increasingly active biomedical sciences field, which  has already standardized vocabularies and ontologies, thus favoring NER initiatives. \par
The construction of parallel corpora of scientific texts has been addressed by different authors.
\newcite{wu2011statistical} constructed a parallel corpus of biomedical article titles from PUBMED in six languages. Based on this corpus, the authors built SMT systems and achieved higher BLEU scores than Google Translator. 
With a different intent, \newcite{kors2015multilingual} produced a gold-standard annotated parallel corpus for biomedical concept recognition in five languages. They used Medline abstracts, drug labels, and patent claims as sources. 
Recently, \newcite{NEVES16.800} used the Scielo scientific database to produce a parallel corpus of biomedical abstracts in three language pairs: Portuguese-English, Spanish-English, and French-English. \par

The Scielo database is a Latin American and Caribbean initiative developed to meet the needs of developing countries regarding scientific communications, increasing the visibility and access to scientific literature \cite{packer2000scielo}. 
Another interesting aspect of Scielo is that several journals publish full-text  of scientific articles in more than one language, a feature commonly limited to the abstracts. Therefore, the Scielo database can be a valuable source for parallel corpora for various scientific domains. \par

In this work, we developed a parallel corpus of full-text scientific articles collected from the Scielo database in English (EN), Portuguese (PT) and, Spanish (ES). 
The corpus is sentence aligned for all language pairs. We also made available trilingual alignments for a subset of sentences. 
The main differences with regard to a previous initiative of \newcite{NEVES16.800} are:  (i) our corpus contains full-text articles, providing a larger resource; (ii) we collected data from various domains, not just biomedical; (iii) whenever possible, we presented the articles in a structured way using sections and paragraphs, favoring other NLP tasks such as summarization; and (iv) we included metadata, such as journal name and subject category, which can be used for text classification.

\section{License and Availability}
Most articles in the Scielo database are licensed under the Creative Commons copyright, with different types of licenses. In order to be able to distribute the contents of the gathered articles, we filtered only those licensed under terms that allow derivatives, since ND (No Derivatives) licenses require the content to be distributed without any modification. As we removed some parts from the articles (e.g. images, tables, references), we would be infringing such copyright rules.
All articles distributed in our dataset contain the corresponding license, authorship, and unique identifiers of original sources, as detailed in Section \ref{sec-statistics}

\section{Material and Methods}
\begin{figure}[!h]
\begin{center}
\includegraphics[scale=0.65]{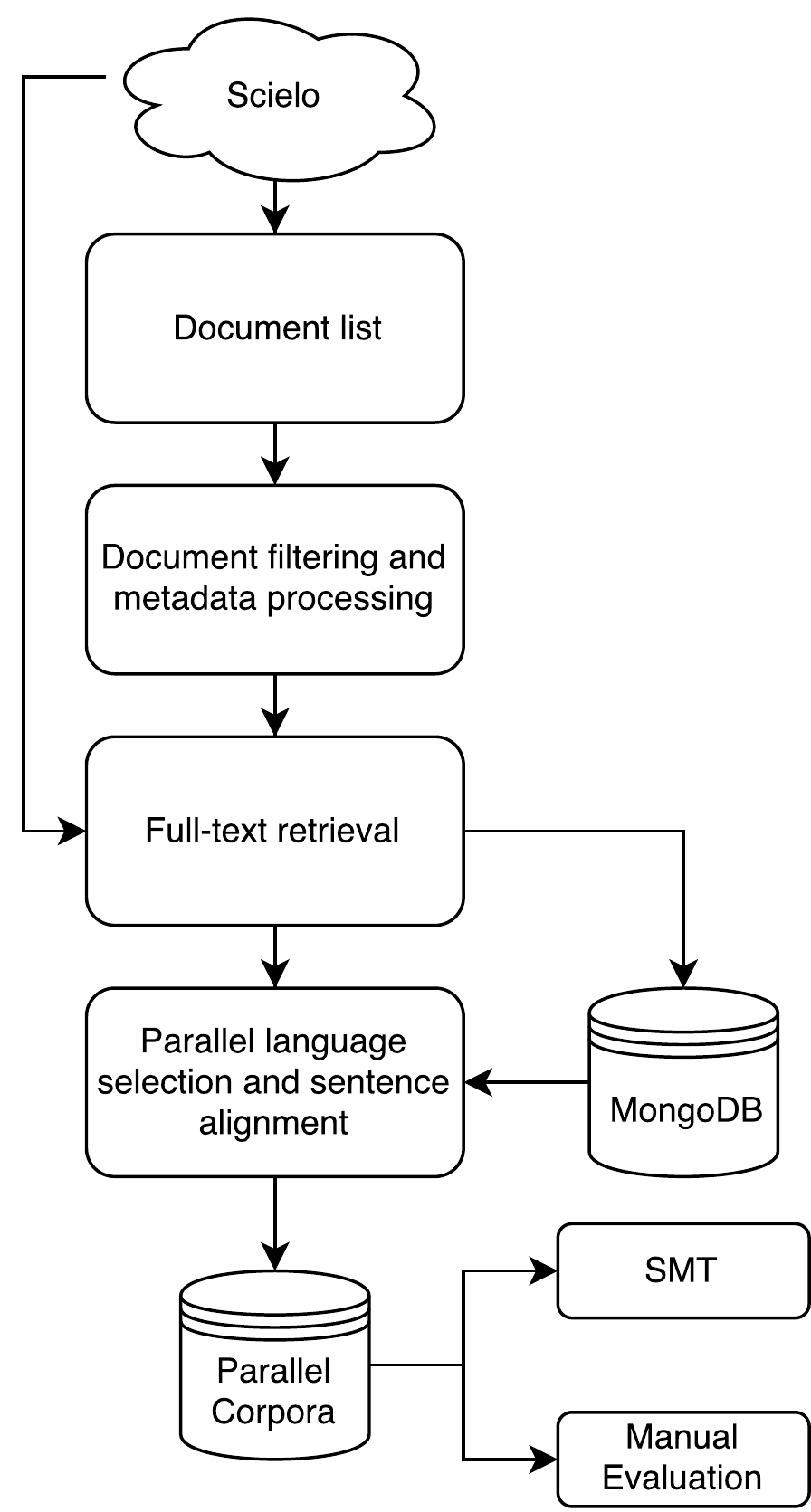} 
\caption{Steps employed in the development of the parallel corpora.}
\label{fig.1}
\end{center}
\end{figure}

In this section, we detail the information gathered from Scielo, the filtering process, as well as our method for article parsing and alignment. An overview of the workflow employed in this article is shown in Figure~\ref{fig.1}

\subsection{Document retrieval}

Scielo's website\footnote{\url{http://www.scielo.org}} provides unified access to a series of regional databases (such as from Argentina, Brazil, South Africa), offering simple and advanced search capabilities. We iteratively queried the database to retrieve all lists of results, which were then parsed and all relevant contents stored, such as URLs for all available languages of each article, authorship, licensing, title, and abstract. We adopted  the MongoDB  database system, as it is  document-oriented, and allows for the easy querying and storage of this type of data. \par


We queried the results in MongoDB to filter only the articles meeting the following constraints: a) articles with full-text available in at least two of three languages of interest (i.e. English, Portuguese, and Spanish); and b)  type of licensing is non ND terms. The full-text of all articles meeting these two criteria were downloaded from the Scielo database in HTML format.

\subsection{Document parsing}
The HTML contents of all articles were parsed using an in-house Python script tailored to the Scielo format. 
First, all non-textual elements, such as images, tables, references, citations, and footnotes were removed. 
Our algorithm was designed to preserve the hierarchical and paragraph structure of the article across the different languages in order to produce results aligned at paragraph and section levels. 
This could help achieving good sentence level alignment. \par

The main challenges in parsing Scielo HTML contents are  heterogeneity  issues concerning HTML structure and formatting over different years. More recent articles are well-formated and contain specific tags for paragraphs, sections, subsections, and titles. 
We concentrated efforts in developing rules to tackle all  ill-formated HTML issues identified, so as to cover as much content as possible, but to reduce the risk of misalignment, we discarded all documents that presented very different structures across the languages. \par

Each parsed full-text translation was stored in MongoDB aiming at preserving the structure of the articles. When our parsing algorithm failed at identifying the document structure, its content was stored as a unstructured list of paragraphs, as we assume that if two translations of the same article present the same number of parsed paragraphs, it is likely they can be simply aligned according to their order. 

\subsection{Sentence alignment}
Once all articles were parsed, we employed a pre-processing step to ensure a better alignment. 
We deleted all parentheses from the texts (mainly used for citations), as well as newline/carriage return characters (i.e \texttt{\textbackslash n} and \texttt{\textbackslash r}), as they would interfere with the sentence alignment tool. \par

For sentence alignment, we used the LF aligner tool\footnote{\url{https://sourceforge.net/projects/aligner/}}, a wrapper around the Hunalign algorithm \cite{varga2007parallel}, which provides an easy to use and complete solution for sentence alignment, including pre-loaded dictionaries for several languages. \par

Hunalign uses Gale-Church sentence-length information to first automatically build a dictionary based on this alignment. Once the dictionary is built, the algorithm realigns the input text in a second iteration, this time combining sentence-length information with the dictionary. When a dictionary is supplied to the algorithm, the first step is skipped. A drawback of Hunalign is that it is not designed to handle large corpora (above 10 thousand sentences), causing large memory consumption. In these cases, the algorithm cuts the large corpus in smaller manageable chunks, which may affect dictionary building.

For articles with the same structure across the languages, pairs of parallel paragraphs were input to the sentence aligner at a time, aiming at reducing the risk of misalignment. For the other cases, all paragraphs were passed to the aligner together. Aligned sentences were stored as text files for post-processing. \par

After sentence alignment, the following post-processing steps were performed: (i) removal of all non-aligned sentences; (ii) removal of all sentences with fewer than three characters, since they are likely to be noise from ill-formatted HTML; 
(iii) removal of all sentences written in the same language using a language detector\footnote{\url{https://github.com/Mimino666/langdetect}}.
This last step was performed since abstracts in different languages could be present in a full-text HTML, which could produce same-language alignments.

\subsection{Machine translation evaluation}
To evaluate the usefulness of our corpus for SMT purposes, we used it to train an automatic translator with Moses \citelanguageresource{moses_lang}. The produced translations were evaluated according to the BLEU score\cite{papineni2002bleu}, using the evaluation tool \texttt{multi-bleu} in Moses.

\subsection{Manual evaluation}
Although the Hunalign algorithm usually presents a good alignment between sentences, we also conducted a manual validation to evaluate the quality of the aligned sentences. We randomly selected 300 pairs of sentences, 100 for each language pair, and 100 trilingual sentences. If the pair was correctly aligned, we marked it as "correct", otherwise, as "no alignment".

\section{Results and Discussion}

In this section, we present statistics about the corpus and a quality evaluation in terms of SMT and sentence alignment. 

\subsection{Corpus statistics} \label{sec-statistics}
Table~\ref{tab:stats} shows the overall corpus statistics for all language pairs and for the set of trilingual aligned documents. One may notice that EN-PT documents are predominant over other language pairs. This may be explained by the fact that almost all Brazilian journals are published through Scielo, thus favoring Portuguese-English translations. \par
The datasets are available\footnote{\url{https://figshare.com/s/091fcaf8ad66a3304e90}} in the TMX format~\cite{Rawat2016}, since it is the standard format for translation memories. Besides the aligned sentences, we included the following metadata for each document: aligned title, authors, copyright license, DOI (if available), journal name, Scielo's unique identifier, and subject area. This information was included either to fully comply with Creative Commons requisites, or to provide additional information for other possible applications, such as text classification or clustering. \par
An example of trilingual sentence is shown below:
%
%

\smallskip
\begin{quote}
English: \textit{Among its objectives, it aims to defend the interests of society and Nursing in the context of Public Policies and the Unified Health System with emphasis on Mental Health.} \par
\smallskip
Spanish: \textit{Entre sus objetivos está defender los intereses de la sociedad y de la Enfermería en el contexto de las Políticas Públicas y del Sistema Único de Salud con énfasis en el área de la Salud Mental.} \par
\smallskip
Portuguse: \textit{Entre seus objetivos, visa defender os interesses da sociedade e da Enfermagem no contexto das Políticas Públicas e do Sistema Único de Saúde com ênfase na área de Saúde Mental.}
\end{quote}

\begin{table}[t]
\centering
\caption{Corpus statistics for all language pairs and the trilingual set. "Docs" refers to the number of documents, "Sents" to the number of aligned sentences, and "Tokens" is the number of tokens in each language.}
\label{tab:stats}
\begin{tabular}{|c|c|c|c|}
\hline
Languages                  & Docs                  & Sents                  & Tokens \\ \hline
\multirow{2}{*}{EN-ES} & \multirow{2}{*}{2,029} & \multirow{2}{*}{177,781} & 5.2M \\ \cline{4-4} 
                   &                    &                    & 5.7M \\ \hline
\multirow{2}{*}{PT-ES} & \multirow{2}{*}{76} & \multirow{2}{*}{4,987} & 140,434 \\ \cline{4-4} 
                   &                    &                    & 151,148 \\ \hline
\multirow{2}{*}{EN-PT} & \multirow{2}{*}{29,609} & \multirow{2}{*}{2.9M} & 76.0M \\ \cline{4-4} 
                   &                    &                    & 77.3M \\ \hline
\multirow{3}{*}{EN-PT-ES} & \multirow{3}{*}{3,142} & \multirow{3}{*}{255,914} & 7.0M \\ \cline{4-4} 
                   &                    &                    & 7.8M \\ 
\cline{4-4} 
                   &                    &                    & 7.2M \\ \hline
\end{tabular}
\end{table}

\subsection{SMT experiments}

Prior to the SMT experiments, all sentences were randomly split in three disjoint datasets for each language pair: training, tuning and test. Approximately 85\% of the aligned sentences were kept for training, 5\% for tuning and 10\% for test. The translation models were built following Moses' baseline system steps\footnote{\url{http://www.statmt.org/moses/?n=moses.baseline}}.

Table~\ref{table-moses} presents the BLEU scores for each language pair for the test set. We included the best results from related work by \newcite{NEVES16.800} for the sake of comparison. We highlight that their study was focused on titles and abstracts from biomedical articles, while our corpus is focused on full-text content of scientific articles in general.\par

\begin{table}[!t]
\centering
\caption{BLEU scores for translation using Moses. Previous related work by Neves et al.(2016) is also presented for comparison in the right-hand column.}
\label{table-moses}
\begin{tabular}{|c|c|c|c|}
\hline
\multicolumn{2}{|c|}{Language Pairs} & \makecell{BLEU \\ Current Work}  		& \makecell{BLEU (Neves \\et al.2016)}\\ \hline
\multirow{2}{*}{EN-ES}  & EN$\rightarrow$ES  & 36.88		& 32.75     \\ \cline{2-4} 
                     	& ES$\rightarrow$EN  & 37.93 		& 30.53    \\ \hline
\multirow{2}{*}{PT-ES}  & PT$\rightarrow$ES  & 62.63 			& -    \\ \cline{2-4} 
                     	& ES$\rightarrow$PT  & 62.96   			& -  \\ \hline
\multirow{2}{*}{EN-PT}  & EN$\rightarrow$PT  & 48.51  		& 33.37   \\ \cline{2-4} 
                     	& PT$\rightarrow$EN  & 49.24   		& 31.78  \\ \hline
\end{tabular}
\end{table}

Our EN-ES models presented considerably higher BLEU scores, despite dealing with several domains. Regarding EN$\rightarrow$ES translation, 
our results are approximately 4.13 percentage points (pp) higher, while about 7.4 pp better for ES$\rightarrow$EN. Considering EN-PT, our results are more expressive. BLEU score for EN$\rightarrow$PT is 15.48 pp higher, and 17.46 pp for PT$\rightarrow$EN. \par
%
%

Another source of comparison is the EuroMatrix project\footnote{\url{http://matrix.statmt.org/matrix}}, which is an European funded initiative to promote machine translation research for all pairs of the European Union (EU) languages. When comparing our results to the reported benchmarking BLEU scores available on the project's website, our results are similar to the reported BLEU scores for the Europarl corpus (scores between 29.4 and 39.4) and overall lower than the ones for the JRC-Acquis corpus (between 55.1 and 60.7). 
The main differences  among these three corpora are domain related, since JRC-Acquis is comprised of EU laws applicable in its state members, Europarl is derived from the EU Parliament proceedings, and our corpus is from scientific articles. Thus, a variation in achieved BLEU scores is expected. As stated by \newcite{kohen_652}, JRC-Acquis corpus is of considerable size within its very specific and well-defined domain of legal text, therefore presenting good translation performance. On the other hand, Europarl corpus is a transcription of speeches, thus inducing a greater linguistic variability.

%


\subsection{Sentence alignment quality}
\begin{figure}[!h]
\begin{center}
\includegraphics[scale=0.65]{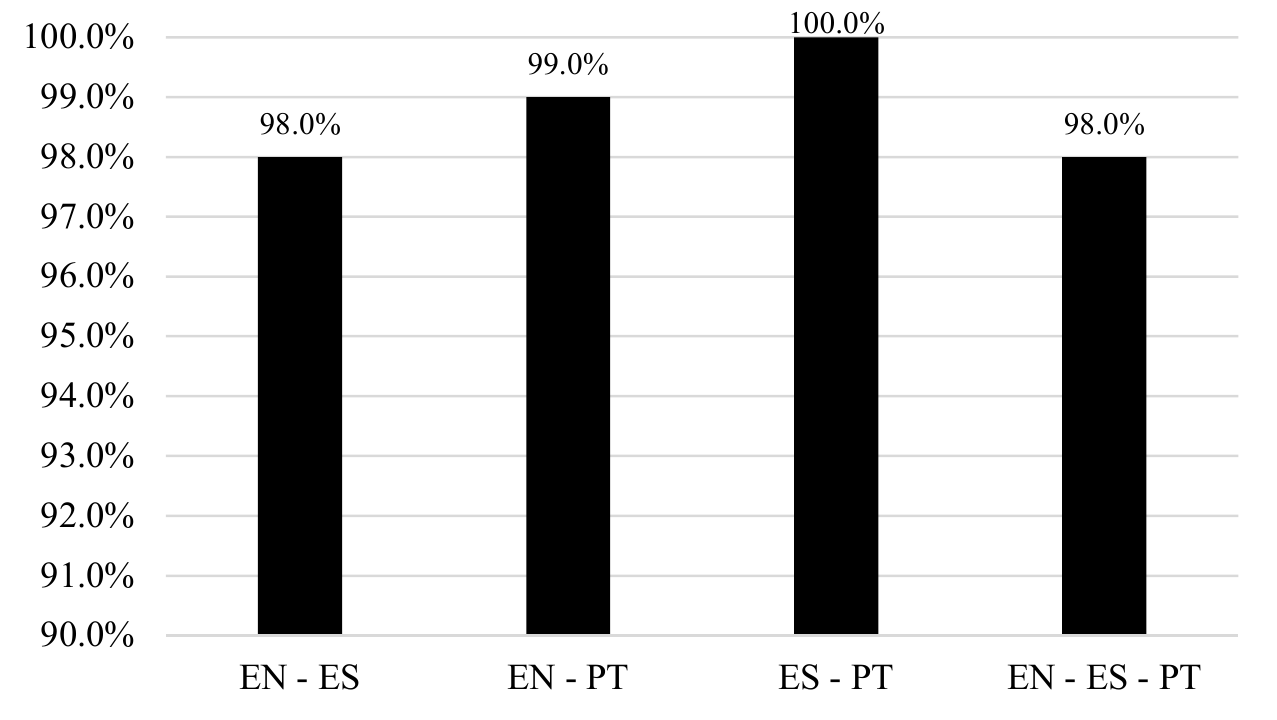} 
\caption{Alignment accuracy for the four language subsets.}
\label{fig.2}
\end{center}
\end{figure}

We manually validated the alignment quality for 400 sentences randomly sampled from the parsed corpus. Figure~\ref{fig.2} depicts the rate of correct alignments for each subset of parallel languages. All language combinations presented at least 98\% of correct alignments, with the language pair ES - PT achieving 100\%.
Different factors may have contributed to this high alignment quality. The use of Hunalign~\cite{varga2007parallel} with a dictionary is perhaps the most probable reason, as it combines a dictionary with sentence-length information to boost alignments. The input of articles segmented by parallel paragraphs also contributed to quality enhancement, since this can reduce the probability of misalignment.

During alignment validation, some translation incoherences were identified, specially regarding nomenclatures, which may have affected the SMT performance. An example of incorrect scientific nomenclature translation is:

\smallskip
\begin{quote}
Portuguese: \textit{[...]movimentos coreoatetosicos nos membros ipsilaterais ao lado comprometido}
\end{quote}
\smallskip
which was translated to:
\smallskip
\begin{quote}
English: \textit{[...] and ipsilateral coreoatetosis}
\end{quote}
\smallskip
while the correct translation would be using "choreoathetosis":
\smallskip
\begin{quote}
English: \textit{[...] and ipsilateral choreoathetosis}.
\end{quote}
\smallskip
Another nomenclature translation problem, but related to context, is:

\smallskip
\begin{quote}
Spanish: \textit{[...]estudiantes en el internado de la Escuela Superior de Medicina}
\end{quote}
\smallskip
which was translated to:
\smallskip
\begin{quote}
English: \textit{[...]boarding school students at the School of Medicine} 
\end{quote}
\smallskip
while a better translation would be:
\smallskip
\begin{quote}
English: \textit{[...]medical intern students at the School of Medicine} 
\end{quote}
\smallskip
since boarding schools are mainly associated to primary and secondary education.

\section{Conclusion and Future Work}

We developed a parallel corpus of scientific articles in three languages: English, Portuguese and Spanish. Additionally to the language pairs, we provided a subset of trilingual aligned sentences. Our corpus is based on full-text contents from the Scielo database, which is available under open-access licenses, thus favoring distribution.

We evaluated our corpus using an SMT experiment with Moses and by manual evaluation of sentence alignment. Our translation experiment presented superior performance regarding BLEU score than a previous related work on Scielo database. We highlight the high translation scores achieved for PT-EN language pairs, boosted by the large number of sentences (almost 3M). Hunalign also presented remarkable alignment quality, with over 98\% sentences correctly aligned. Other important features of our corpus are the availability of trilingual sentences, and the additional subset of articles aligned according their hierarchical structures, which can be useful for automatic building of structured abstracts.

Regarding future work, we foresee the use of this corpus in text mining applications, such as classification and clustering. In addition, the corpus could be used in cross-language plagiarism detection, as we provide at least two versions of the same article in multiple languages. New Neural Machine Translation (NMT) systems could be trained to provide a comparison against the tested SMT system. An interesting approach would be use an approach similar to Google's multilingual NMT \cite{johnson2017google} to perform zero-shot translation based on an additional parallel corpus.

\balance
\section{Bibliographical References}
\label{main:ref}

\bibliographystyle{lrec}
\bibliography{xample}

\section{Language Resource References}
\label{lr:ref}
\bibliographystylelanguageresource{lrec}
\bibliographylanguageresource{xample}

\end{document}